\documentclass{article}

\usepackage[final]{nips_2017}

\usepackage[utf8]{inputenc} 
\usepackage[T1]{fontenc}    
\usepackage{hyperref}       
\usepackage{url}            
\usepackage{booktabs}       
\usepackage{amsfonts}       
\usepackage{nicefrac}       
\usepackage{microtype}      

\title{The Doctor Just Won't Accept That!}

\author{
  Zachary C. Lipton\\
  Carnegie Mellon University,\\
  Amazon AI\\
  \texttt{zlipton@cmu.edu}
}

\begin{document}
\maketitle

\emph{``I work with medical data. 
We work with doctors
and they're interested in 
predicting risk of mortality, 
recognizing cancer in radiologic scans,
and spotting diagnoses based on 
electronic health record data.
We can train a model,
and it can even give us the right answer.
But we can't just tell the doctor
``my neural network says this patient has cancer!''
The doctor just won't accept that!
They want to know why the neural network says what it says. 
They want an explanation.
They need interpretable models."\\
--- Prototypical call to arms 
} 

\section*{Introduction}


Each day, machine learning practitioners integrate predictive models 
more deeply into real-world decisions. 
As the steady creep touches
sensitive domains, like medical decision-making, financial lending, and criminal justice,
refrains like the fictional exemplar printed above 
echo throughout technical, industrial, and political spheres.
In order to know if the model is 
\emph{\{fair, safe, reliable, robust, sensible\}}, we need to know \emph{how} the model makes its decisions. 
Or perhaps we need to know \emph{why} the model makes its decisions. 
We will return to interpretability's polysemy \citep{lipton2016mythos} later. 


These calls to arms 
express a well-founded discomfort 
with machine learning.
How can a software agent that  
does not even know \emph{what} a loan is
decide \emph{who} qualifies for one?
Indeed, we ought to be cautious about injecting machine learning (or anything else, for that matter) into applications where there may be a significant risk of causing social harm.

However, \emph{``the doctor just won't accept that''} does not provide a solid foundation  for a proposed field of study. 
For the field of interpretable machine learning to advance,
we must ask the following questions:
\emph{What precisely won't the various stakeholders accept?
What do they want?
Are these desiderata reasonable?
Are they feasible?}
Returning to automated lending, 
if stakeholders demanded that the decision-maker understand what a loan is,
and what it means to pay one off,
and the purpose of an explanation 
were to demonstrate such understanding,
\emph{then no conceivable interpretation technique could salvage a supervised learning model}.
In order to answer these questions, 
we'll have to give real-world problems 
and their respective stakeholders 
greater consideration.

\section*{The Interpreters}

Eager to heed the demand for interpretable models, 
researchers regularly propose 
algorithms claimed to constitute \emph{interpretable} machine learning.
However, these papers
wield the term \emph{interpretability}
in many disparate senses.
For example, some propose that interpretable models are by definition \emph{simple},
tying the interpretability of the model to  the (inverse) cognitive strain endured by a would-be interpreter.
Others advocate models where each parameter has some intuitive significance.
For example, linear models are often held up as interpretable owing to the correspondence between each weight and an input feature.
Practitioners and customers might 
eyeball the parameters to see whether they accord with intuition.
While this mode of interpreting 
linear models may be specious,
we put off that criticism for now.
Some advocate models with monotonic relationships between inputs and outputs.
Others suggest post-hoc techniques for explaining models.

However, amid the worry of machine learning run amok,
the ensuing call for interpretability,
and the rush to satisfy this demand,
technical solutions have arrived ahead of technical questions. 
Papers claim interpretability,
often axiomatically, without defining it. 
Sometimes the proof that a model is interpretable is simply that it yields an appealing visualization. Just look, see how interpretable it is!

Confusing matters,
the conversation around interpretability
has lurched to the fore
amid the rise of deep learning.
As a result, some specific interpretability problems
are misattributed to the opacity of neural networks 
even when they are shared more widely by all supervised learning algorithms. 
For example, linear models also fail to differentiate essential vs. spurious correlations. 
They do not necessarily elucidate causal structure
and they are not cognizant of human notions of fairness. 

In the rush to satisfy demand, important steps are routinely skipped. Namely, we fail to ask what end the proposed interpretability serves. We also fail to ask if these desiderata can be satisfied by new models within current frameworks (say, supervised learning) or if they speak to fundamental limitations of the paradigm. Only when we know what is desired and that it might feasibly be provided within a given paradigm,
does it make sense to propose model properties and post-hoc techniques.
Absent these considerations, algorithmic proposals lack clear problem definitions.
With a surfeit of hammers, 
and no agreed-upon nails,
the field risks failing to mature 
owing to shaky foundations.

\section*{The Detractors}

Meanwhile, detractors often question 
the entire enterprise of interpretable machine learning,
suggesting that the calls to arms 
are misguided.
Others question whether human decisions 
are interpretable in the first place. 
The logic goes: 
if we presently trust humans with all decisions, 
despite their lacking interpretability, 
then why should interpretability be required
of the decision-makers that succeed us?
However, these ``how do we know that humans  can X?'' questions sit on shaky ground as well. 
Every speaker at this conference can explain why they submitted papers, and even elucidate a precise chain of events by which they expected the action might culminate in a trip to California. 
Surely, humans can explain their actions 
by reference to a robust knowledge 
of the dynamics by which our society and the physical world operate, 
and surely such reasoning speaks to what \emph{some (but not all)} 
of what people mean by \emph{explanations}.

To drive the point home, 
of all the people who claim 
that humans are unable to explain their actions,
it would be hard to find one person among them who actually lives by this logic, 
never asking any other person 
to explain their actions.
While humans can explain actions,
they are neither transparent, parsimonious (in parameters), decomposable, nor monotonic.
Humans may lie, and our explanations may fail to fully explain our decisions, 
but the whole sale dismissal of human explainability tramples on common sense.

Both proponents and detractors of interpretable machine learning go astray in similar fashion:
by failing to fix more specific definitions.
Is a linear model interpretable? 
In the sense that it is simple, \emph{yes}.
In the sense that it provides verbal explanations of its actions, \emph{maybe, via  post-hoc techniques}.
In the sense that it explains what dynamics in the world it depends upon, decidedly not.
For a human decision-maker, 
the answers are nearly opposite. 
In order to advance our knowledge of 
machine learning and its amenability to interpretation,
we must move past the word interpretable and 
decide which definitions to focus on.
Moreover, we must confront
the normative question of deciding which ones are important.

\section*{The Stakeholders}

If demands for interpretability speak to real shortcomings of machine learning, especially vis-a-vis socially impactful applications,
and if interpretable machine learning simultaneously admits no clear definition 
that might serve as a foundation 
upon which a traditional technical 
(vs. critical) discipline might flourish,
then what next steps might we take?
What should we do at an interpretability symposium?
Should we teach beginners the current tools of interpretable AI?
Or should we debate the critical questions,
sorting out which problems demand a new model and which demand categorically different forms of artificial intelligence?

Curiously, while interpretable machine learning 
is touted to address the demands of stakeholders,
those stakeholders are seldom 
taken seriously.
In machine learning papers, 
we consider their needs carelessly.
While variants of ``the doctors just won't accept that!'' echo throughout convention centers packed with machine learners across the globe, 
few have bothered to ask 
what precisely they do want. 

In my subjective experience,
the policy communities take a considerably broader view. 
At NYU Law, a recent workshop on Algorithms and Explanations \citep{nyu2017algos} 
engaged policy experts alongside stakeholders in fields spanning medicine, journalism, policing, and lending, and also machine learning scientists.

By contrast, the machine learning community, despite claiming these topics as motivation, remains conspicuously isolated.
More mature subfields might thrive amid such disciplinary modularity.
Image recognition is sufficiently advanced
and its applicability to industrial problems so well understood, 
that you could develop a classifier for ripe vs. unripe fruit 
without a deep investment in the produce industry.
The API is known by both stakeholders and scientists.
But absent consensus on problem definitions,
this isolation can impede progress.
Consider the widely-discussed provision of the European Union General Data Privacy Regulation (GDPR) popularly described as a \emph{right to explanation}. 
While both the meaning and enforceability of the law and whether it amounts at all to a \emph{right to explanation} remain under debate \citep{wachter2017right,goodman2016european},
machine learning papers reference the ordinance glancingly via the one paper \citep{goodman2016european} which presented the material for a machine learning audience.
Similar issues have emerged in algorithmic fairness, 
where a single genre-crossing paper \citep{barocas2016big} holds uncontested authority owing not only its high quality, but also to a general ignorance of policy in machine learning circles. 

\section*{The Conclusions}

These are hard questions and the academic incentives don't encourage researchers to pursue them. 
Careers are judged on technical work,
and technical machine learning conferences do not frequently publish position papers.
So we find ourselves in a world 
where machine learning conferences will publish papers 
that offer a claimed interpretable algorithm,
but not those that address the foundations of such claims.
For work in this field to be meaningful 
and for this field to progress, 
we must take problem formulation more seriously. 
The conversation must move beyond 
``the doctor just won't accept that!''
We should determine what the various stakeholders demanding interpretability want, 
and which of these desiderata 
can actually be satisfied 
within the current learning paradigm. 
To do any of this effectively, 
we must invite the stakeholders 
to participate in the conversation.

\bibliographystyle{plainnat}
\bibliography{symposium.bib}
\end{document}